%% file: iclr2023_conference_tinypaper.tex
\documentclass{article} 
\usepackage{iclr2023_conference_tinypaper,times}

\input{math_commands.tex}

\usepackage{hyperref}
\usepackage{url}
\usepackage{graphicx}

\title{MosquitoFusion: A Multiclass Dataset for Real-Time Detection of Mosquitoes, Swarms, and Breeding Sites Using Deep Learning}

\author{Md. Faiyaz Abdullah Sayeedi, Fahim Hafiz \& Md Ashiqur Rahman\\
Department of Computer Science and Engineering, United International University\\
{\small\texttt{msayeedi212049@bscse.uiu.ac.bd,\{fahimhafiz,ashiqurrahman\}@cse.uiu.ac.bd}} \\
}

\iclrfinalcopy 
\begin{document}

\maketitle

\begin{abstract}
In this paper, we present an integrated approach to real-time mosquito detection using our multiclass dataset (MosquitoFusion) containing 1204 diverse images and leverage cutting-edge technologies, specifically computer vision, to automate the identification of Mosquitoes, Swarms, and Breeding Sites. The pre-trained YOLOv8 model, trained on this dataset, achieved a mean Average Precision (mAP@50) of 57.1\%, with precision at 73.4\% and recall at 50.5\%. The integration of Geographic Information Systems (GIS) further enriches the depth of our analysis, providing valuable insights into spatial patterns. The dataset and code are available at \url{https://github.com/faiyazabdullah/MosquitoFusion}. 



\end{abstract}

\section{Introduction}

Mosquito-borne diseases stand as a major global health threat due to the adaptability and resilience of mosquitoes. Roughly 700 million people are infected with mosquito-borne diseases every year. \cite{Adnan} An estimated 1 million people die from these diseases annually. Combatting these diseases requires a reevaluation of existing strategies. Understanding mosquito breeding grounds and behaviors is crucial for effective prevention. This research tackles the broader challenge of preventing mosquito-borne diseases by emphasizing the swift detection of mosquitoes. In this paper, we contribute to this effort by leveraging our extensive MosquitoFusion dataset. To understand the usability of our dataset, we implement the pre-trained YOLOv8 object detection model.

\section{Existing Works}

Recent research has witnessed a surge in multidisciplinary approaches to combat mosquito-borne diseases. \cite{R1} pioneered an automated detection system using unmanned aerial vehicles (UAVs) for identifying potential breeding sites, emphasizing the efficacy of aerial surveillance. \cite{R2} leveraged Convolutional Neural Networks (CNNs) and geospatial analysis, highlighting the synergy between advanced algorithms and geographic insights. \cite{R4} integrated Geographic Information Systems (GIS) for refined risk assessments, emphasizing the correlation between breeding sites and environmental factors. Recent studies by \cite{R6} and \cite{R7} underscore the significance of image annotation precision and preprocessing techniques for improved model accuracy. However, existing works focused on creating the datasets in laboratory environments \cite{labenv} and often lack multi-class diversity. Such datasets of mosquitoes do not consider real-life aspects in images. Also, many datasets concentrate on a single aspect, like mosquito \cite{mosData} or breeding site \cite{breedData} detection. Moreover, a majority of these datasets are not publicly accessible. 


\begin{figure}[t]
    \centering
    \includegraphics[width=\textwidth, height=0.3\textheight]{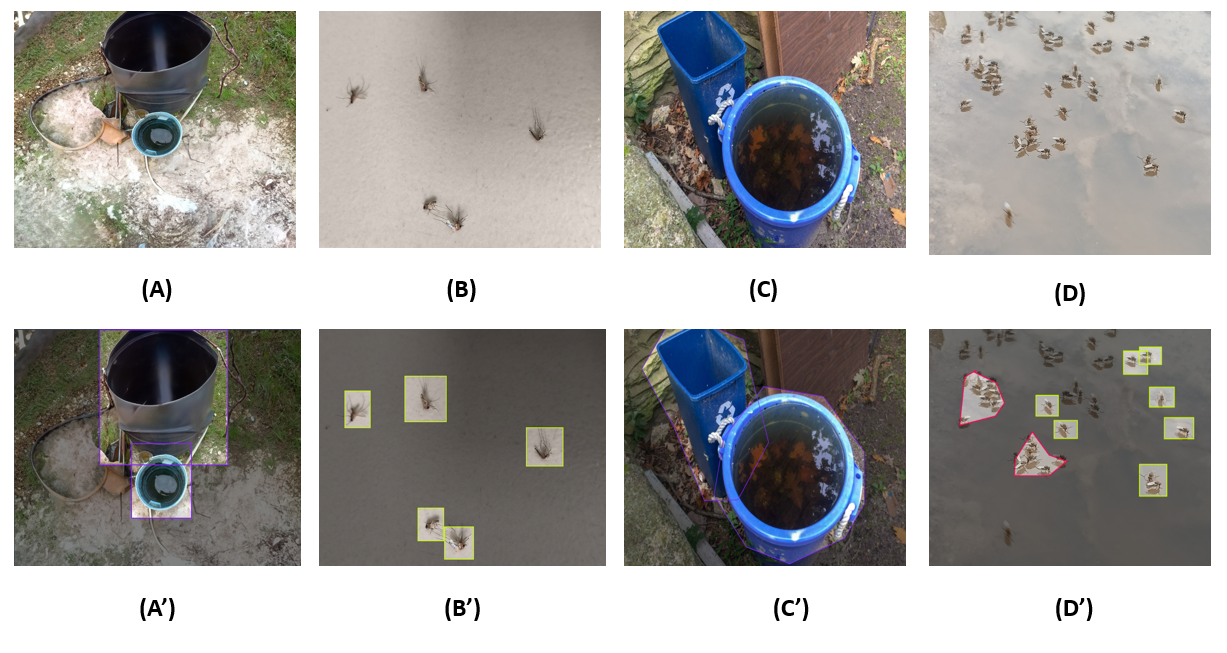}
    \caption{Fully tagged and labeled image. (A), (B), (C), (D) shows the original image. In (A') and (C'), the purple borders mark the breeding sites. In (B') and (D') the yellow borders mark the mosquitos. (D') The red borders mark the swarms} 
    \label{fig:annotate}
\end{figure}

\section{Methodology}

In Fig. \ref{fig:annotate} we present an overview of the annotated dataset. In section \ref{dataset} and \ref{model} we present our dataset and technical validation.

\subsection{Dataset}
\label{dataset}
The dataset comprises 1204 meticulously curated images, strategically divided into training (87\%), validation (8\%), and test (5\%) sets, totaling 1053, 100, and 51 images, respectively. Rigorous preprocessing \ref{A2} measures ensure high-quality data. Augmentations such as flips, rotations, crops, and grayscale applications enhance dataset diversity. This meticulously prepared dataset serves as a valuable resource for training, validating, and testing models for mosquito detection. More details of the dataset have been discussed in the section \ref{appendix}. In figure \ref{fig:methodology} we present the total overview of our methodology.



\subsection{Technical Validation}
\label{model}
We use the pre-trained 'YOLOv8s' object detection model \cite{yolov8_ultralytics} which utilizes the CNN architecture, to evaluate our dataset. This configuration achieved a mean Average Precision (mAP@50) of 57.1\%, with a precision of 73.4\% and a recall of 50.5\%. This configuration aligns with the objective of efficient and accurate mosquito identification. Then we integrate the Geographic Information Systems (GIS) to further enrich the depth of our analysis, providing valuable insights into spatial patterns. The summary of evolution matrices is shown in Table \ref{tab:evaluation-metrics}. More technical validation is shown in \ref{A4}.

\section{Result Analysis and Future Work}
The model trained on the MosquitoFusion dataset exhibits promising performance, showcasing its efficacy in real-time mosquito detection. The dataset's careful curation and diverse augmentations contribute to the model's robustness. The split into training, validation, and test sets ensures reliable evaluation, emphasizing the dataset's value for training effective mosquito detection model. Beyond its utility in research, the dataset holds great potential for applications in public health, environmental monitoring, and disease control strategies. Our future work includes creating a custom model exclusively designed for detecting mosquitoes, swarms, and breeding sites to further advance our capabilities in this domain. Additionally, we'll address a limitation of the pre-trained YOLOv8 model in our future work that, it may struggle to differentiate swarms formed by mosquitoes from those formed by other insects.

\subsubsection*{URM Statement}
The authors acknowledge that at least one key author of this work meets the URM criteria of ICLR 2024 Tiny Papers Track.

\bibliography{iclr2023_conference_tinypaper}
\bibliographystyle{iclr2023_conference_tinypaper}

\appendix
\section{Appendix}
\label{appendix}
In sections \ref{A1}, \ref{A2}, \ref{A3}, and \ref{A4}, we present our data collection, data preprocessing, distribution analysis and folder structure, and the model setup and evaluation of the dataset.

\subsection{Data Collection}
\label{A1}
In the initial phase of our project, data collection for the MosquitoFusion dataset involved meticulous fieldwork, employing professional cameras to capture 1204 detailed images of mosquitoes, swarms, and breeding sites. We have captured the images in the lighting conditions of daylight and in a sunny environment. The dataset's reliability is underscored by careful annotation using the tool Roboflow \cite{roboflow}. This hands-on approach ensures the acquisition of authentic and representative images for effective real-time mosquito detection.

\subsection{Data Preprocessing}
\label{A2}

Data preprocessing is a crucial step in ensuring the quality and effectiveness of a dataset for training machine learning models. In the case of the MosquitoFusion dataset, our preprocessing pipeline includes several key steps, starting with data cleaning and curation. Then we did auto-orientation and resizing to a consistent 640x640 pixel dimension. We also filter out images lacking annotations for integrity. Then we did augmentations, including flips, rotations, crops, and grayscale applications to introduce variability. With a total of 1204 images strategically divided into training, validation, and test sets, the preprocessing emphasizes creating a standardized yet diverse dataset.

\subsection{Distribution Analysis and Folder Structure}
\label{A3}

The dataset encompasses instances of three distinct classes: Breeding Place, Mosquito, and Mosquito Swarm. Specifically, the Breeding Place Class is represented by 1031 instances, the Mosquito Class includes 133 instances, and the Mosquito Swarm Class comprises 40 instances. In table \ref{tab:class-distribution}, we present the class distribution of our dataset.

The dataset appears imbalanced because capturing images of mosquitoes and swarms is quite challenging. Unlike other objects, mosquitoes are small, swift, and often found in dynamic swarms, making it harder to obtain clear images. To tackle the imbalance in the dataset, we employ the technique called oversampling. This involves increasing the number of instances for the imbalanced classes by using data augmentation methods. This helps ensure that the model is exposed to a more balanced representation of all classes, enhancing its ability to recognize instances from each category effectively.

\begin{figure}[h]
    \centering
    \includegraphics[width=\textwidth, height=9.5cm]{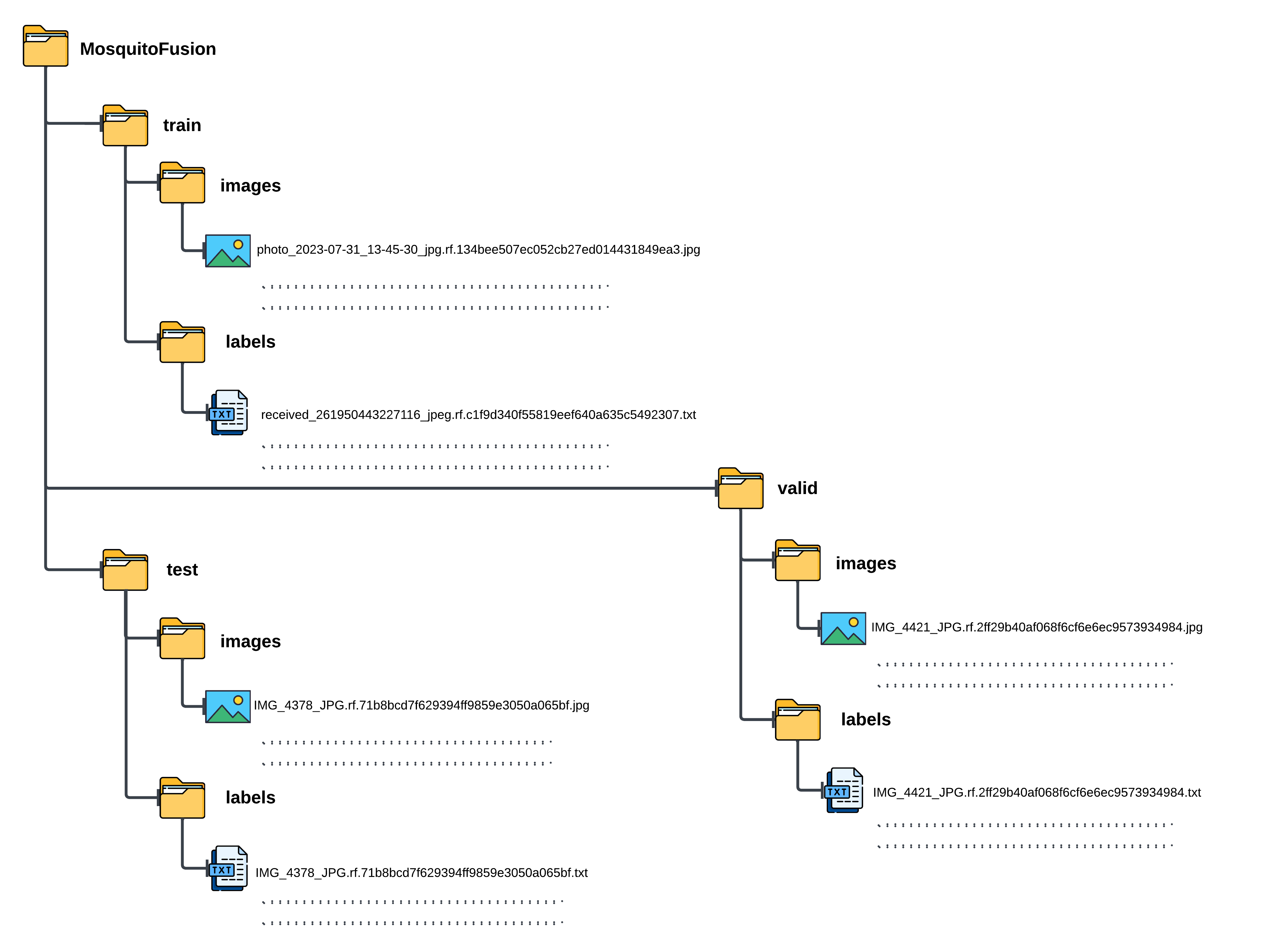}
    \caption{The folder structure of the MosquitoFusion dataset}
    \label{fig:dataset}
\end{figure}

Within each directory - Train, Valid, and Test - two folders, namely "image" and "label" organize the dataset. This dual-folder structure streamlines data management, with the "image" folder housing the visual representations, and the "label" folder containing corresponding annotations. This meticulous organization enhances the dataset's usability. In figure \ref{fig:dataset} we present the folder structure of our dataset.

\begin{table}[h]
    \centering
    \small
    \begin{tabular}{lccc}
        \hline
        \textbf{Class} & \textbf{Instances} \\
        \hline
        Breeding Place & 1031 \\
        Mosquito & 133 \\
        Mosquito Swarm & 40 \\
        \hline
        \textbf{Total} & 1204 \\
        \hline
    \end{tabular}
    \caption{Distribution of Classes in the Dataset}
    \label{tab:class-distribution}
\end{table}



\subsection{Model Setup and Evaluation}
\label{A4}
All the images in the dataset were manually reviewed to ensure that no individually identifiable information was included or embedded in the dataset. To make sure the dataset is appropriate for training deep learning models we trained the localization model using the pre-trained YOLOv8s model. The images were randomly split into 87\% (1053) training, 8\% (100) validation, and 5\% (51) test images for training and testing the localization model.

The training process took place on a Windows 11 (Version 23H2) machine running, equipped with Nvidia RTX 3070Ti GPU boasting 8GB of video memory and an AMD Ryzen 5800X processor. The model underwent pre-training using the COCO3, running for a total of 25 epochs. The input size was set to 640 pixels, and standard hyperparameters were employed throughout the training sessions.


\textbf{Object Detection Performance: }For the localization task, the Mosquitos, Swarms, and Breeding Sites were detected with a box precision of 73.4\%, recall of 50.5\%, and mean Average Precision (mAP@50) of 57.1\% at IoU of the 50th percentile on the validation set.

\begin{table}[h]
    \centering
    \small
    \begin{tabular}{|l|l|}
    \hline
    \textbf{Model Type} & YOLOv8s \\
    \hline
    \textbf{Architecture} & CNN \\
    \hline
    \textbf{mAP@50} & 57.1\% \\
    \hline
    \textbf{Precision} & 73.4\% \\
    \hline
    \textbf{Recall} & 50.5\% \\
    \hline
    \end{tabular}
    \caption{Evaluation Metrics}
    \label{tab:evaluation-metrics}
\end{table}

\begin{figure}[h]
    \centering
    \includegraphics[width=\textwidth, height=0.22\textheight]{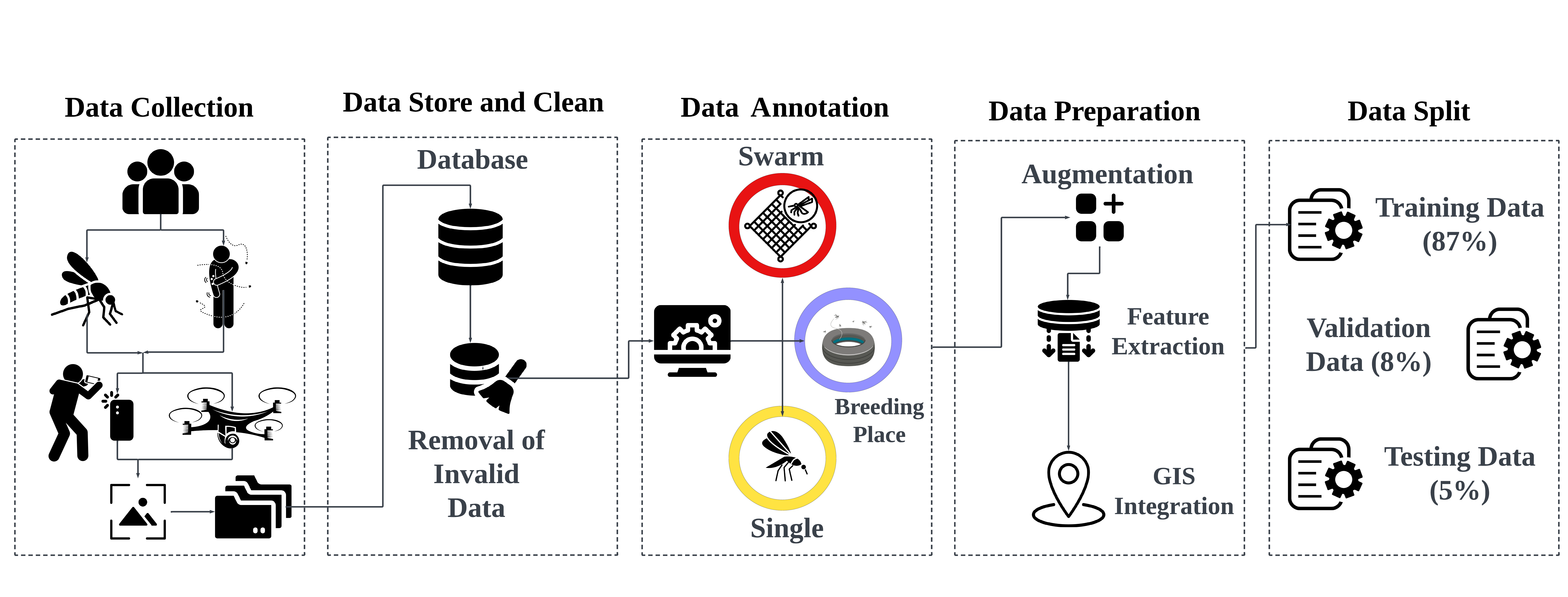}
    \caption{Framework of Methodology} 
    \label{fig:methodology}
\end{figure}

\end{document}

%% file: math_commands.tex
\usepackage{amsmath,amsfonts,bm}

\def\eqref#1{equation~\ref{#1}}

\def\1{\bm{1}}

\DeclareMathAlphabet{\mathsfit}{\encodingdefault}{\sfdefault}{m}{sl}
\SetMathAlphabet{\mathsfit}{bold}{\encodingdefault}{\sfdefault}{bx}{n}

